\documentclass[11pt]{article}

\usepackage[preprint]{acl}

\usepackage{times}
\usepackage{latexsym}

\usepackage[T1]{fontenc}
\usepackage[utf8]{inputenc}

\usepackage{microtype}

\usepackage{inconsolata}

\usepackage{graphicx}

\usepackage{hyperref}       % hyperlinks
\usepackage{url}            % simple URL typesetting
\usepackage{booktabs}       % professional-quality tables
\usepackage{amsfonts}       % blackboard math symbols
\usepackage{nicefrac}       % compact symbols for 1/2, etc.
\usepackage[table]{xcolor}
\usepackage{graphicx}
\usepackage{multirow}
\usepackage{makecell}
\usepackage{pifont}       
\usepackage{bbding}      
\usepackage{fontawesome}
\usepackage{subcaption}
\usepackage{arydshln}
\usepackage{amsmath}
\usepackage{tabularx}
\usepackage{booktabs}
\usepackage{mdframed}
\usepackage[most]{tcolorbox}
\usepackage{arydshln}
\usepackage{adjustbox}

\definecolor{softred}{RGB}{251,231,229} 
\definecolor{table-blue}{RGB}{173, 216, 230}
\definecolor{table-yellow}{RGB}{255,245,233}
\newtcbox{\thinktag}{
  on line,
  colback=blue!10,
  colframe=blue!10,
  boxrule=0pt,
  arc=5pt,
  boxsep=0.8pt,
  left=0pt, right=0pt,
  top=0pt, bottom=0pt,
  nobeforeafter,
  valign=center,
  fontupper=\ttfamily\color{blue!70!black},
}

\newtcbox{\informationtag}{
  on line,
  colback=orange!10,
  colframe=orange!10,
  boxrule=0pt,
  arc=5pt,
  boxsep=0.8pt,
  left=0pt, right=0pt,
  top=0pt, bottom=0pt,
  nobeforeafter,
  valign=center,
  fontupper=\ttfamily\color{orange!60!black},
}

\newtcbox{\routetag}{
  on line,
  colback=green!10,
  colframe=green!10,
  boxrule=0pt,
  arc=5pt,
  boxsep=0.8pt,
  left=0pt, right=0pt,
  top=0pt, bottom=0pt,
  nobeforeafter,
  valign=center,
  fontupper=\ttfamily\color{green!60!black},
}

\newtcbox{\answertag}{
  on line,
  colback=red!5,
  colframe=red!5,
  boxrule=0pt,
  arc=5pt,
  boxsep=0.8pt,
  left=0pt, right=0pt,
  top=0pt, bottom=0pt,
  nobeforeafter,
  valign=center,
  fontupper=\ttfamily\color{red!70!black},
}

\newcommand{\think}[1]{\thinktag{<think>}~#1~\thinktag{</think>}}
\newcommand{\info}[1]{\informationtag{<information>}~#1~\informationtag{</information>}}
\newcommand{\route}[1]{\routetag{<search>}~#1~\routetag{</search>}}
\newcommand{\answer}[1]{\answertag{<answer>}~#1~\answertag{</answer>}}

\title{ReCal: Reward Calibration for RL-based LLM Routing}

\author{
  \textbf{Qihang Yu\textsuperscript{1}},
  \textbf{Hanwen Tong\textsuperscript{2}},
  \textbf{Zhengqi Zhang\textsuperscript{2}},
  \textbf{Bo Zheng\textsuperscript{2}},
  \\
  \textbf{Feng Wei\textsuperscript{2}},
  \textbf{Shengyu Zhang\textsuperscript{1,*}},
  \textbf{Zemin Liu\textsuperscript{1,*}},
  \textbf{Fei Wu\textsuperscript{1,3}}
  \\
  \\
  \textsubscript{1}Zhejiang University,
  \textsubscript{2}Ant Group,
  \textsubscript{3}Shanghai AI Laboratory
\\
\texttt{
\{yuqihang, sy\_zhang, liu.zemin, wufei\}@zju.edu.cn,
}\\
\texttt{
\{tonghanwen.thw, zzq513183, huodeng.wf\}@mybank.cn,zhengbo\_321@163.com
}\\
$^{*}$ Corresponding author
}

\begin{document}
\maketitle
\begin{abstract}
Large language model (LLM) routing has emerged as an effective paradigm for leveraging the complementary strengths of multiple LLMs through dynamic model and reasoning-strategy selection. Recent reinforcement learning (RL)-based routing methods further improve routing quality by optimizing routing policies from interaction feedback. However, they still struggle to provide informative and comparable learning signals under heterogeneous tasks with varying difficulty.
In practice, multiple objectives (e.g., correctness, format behavior) are aggregated into a single scalar reward, leading to ambiguous credit assignment and conflicting optimization signals. Moreover, reward signals exhibit significant variability across instances, where some instances produce higher or more variable rewards, introducing optimization bias that favors trivial samples over informative ones.
To address these issues, we propose \textbf{ReCal}, a \textbf{\underline{Re}}ward \textbf{\underline{Cal}}ibration framework for RL-based LLM routing. We first introduce a hierarchical reward decomposition mechanism with component-wise advantage estimation.
We further propose a distribution-aware optimization strategy that calibrates optimization variability
through variance-aware reweighting and per-dataset normalization.
Experiments on seven datasets demonstrate that ReCal consistently improves routing performance, and training stability over baselines\footnote{Code: \url{https://anonymous.4open.science/r/ReCal}}.
\end{abstract}

\section{Introduction}
\begin{figure*}[ht]
    \centering
    \includegraphics[width=\linewidth]{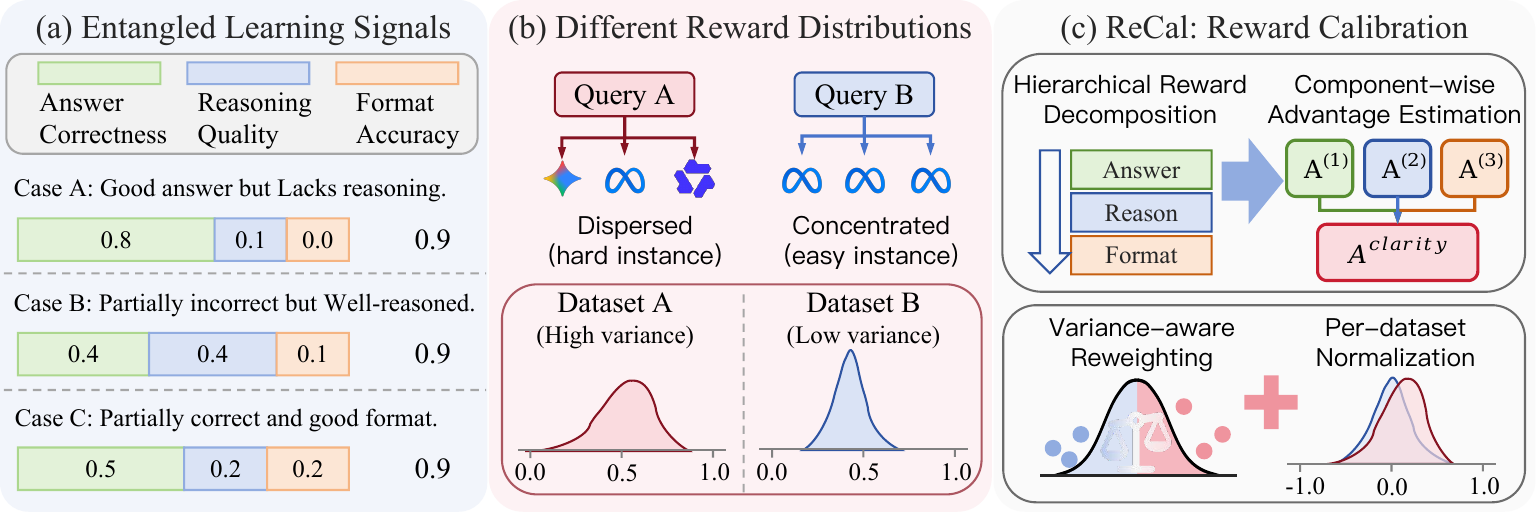}
    \caption{(a) Scalarized rewards entangle multiple objectives, leading to ambiguous credit assignment. (b) Reward distributions vary widely across routing instances and datasets, causing optimization to be dominated by high-variance or easier cases, while informative cases are under-optimized. (c) ReCal provides clear, component-wise learning signals and applies distribution-aware optimization to calibrate policy updates.}
    \label{fig:intro}
\end{figure*}
LLM routing has emerged as an effective paradigm for leveraging the complementary strengths of multiple LLMs by dynamically selecting appropriate models or reasoning strategy for each query. Among existing approaches, a line of recent work formulates routing as a sequential decision-making problem and applies RL to optimize routing policies~\cite{sikeridis2025pickllm,Router-R1,qian2025xrouter,chen2005qos}, achieving promising performance gains.
Despite recent progress, existing routing methods often struggle to learn effective policies under queries with varying difficulty and routing uncertainty. In practice, learning signals can exhibit uneven statistical properties across instances and datasets, leading to inconsistent policy updates during training and biased optimization toward easier decision patterns. Consequently, routing policies may fail to learn reliable behaviors for challenging cases.

We argue that a key bottleneck of RL-based LLM routing lies in the learning signals used for routing policy optimization. Unlike standard generation tasks, routing policies must learn query-dependent model-selection behaviors under heterogeneous routing difficulty. However, there are two key technical issues with learning signals in current routing frameworks.
First, routing policies are typically optimized using scalarized rewards that aggregate multiple objectives, such as answer correctness and formatting behavior, which often correspond to different routing decisions and model-selection strategies into a single score. As shown in Figure~\ref{fig:intro}(a), similar reward improvements may arise from entirely different routing behaviors, making it difficult for the policy to identify which routing decision performs better. Consequently, scalarized rewards obscure routing credit assignment and weaken policy learning~\cite{MANNION201760,SILVER2021103535,huang2021reward}.
Second, routing rewards exhibit highly heterogeneous statistical properties across instances and datasets. As shown in Figure~\ref{fig:intro}(b), some routing cases produce highly dispersed rewards due to large disagreement among candidate models, while others yield concentrated rewards with limited routing ambiguity. Different datasets further induce substantially different reward scales and variances. Such heterogeneity can bias optimization toward easier routing cases~\cite{schulman2017ppo,yu2025reward}, while informative but underrepresented routing decisions receive insufficient learning attention.
These observations raise two central challenges for RL-based LLM routing: \textbf{(1)} how to transform entangled routing rewards into clear learning signals for precise routing credit assignment; and \textbf{(2)} how to calibrate optimization signals across heterogeneous routing instances and datasets, thereby assigning appropriate emphasis to uncertain and informative routing decisions during optimization.

To address these challenges, we propose a \textbf{\underline{Re}}ward \textbf{\underline{Cal}}ibration Framework for LLM routing, abbreviated as \textbf{ReCal}. Rather than directly optimizing with raw routing rewards, ReCal treats policy learning as a two-stage calibration process: it first restructures reward signals into clearer supervision, and then adjusts the contribution of these signals to policy updates under heterogeneous data distributions.
We introduce a hierarchical, disentangled reward mechanism to improve learning signal clarity \textbf{(Challenge 1)}.  Instead of estimating a single advantage from aggregated scalarized rewards, we separately estimate the advantage of each reward component, which preserves objective-specific relative improvements during optimization. The resulting component-wise advantages are then combined through weighted aggregation to reflect the relative importance of different objectives while reducing interference among conflicting optimization signals.
Building on the calibrated rewards, we further improve signal comparability \textbf{(Challenge 2)} under heterogeneous reward distributions through a distribution-aware optimization strategy. At the instance level, we introduce variance-aware reweighting to dynamically calibrate update strength according to disagreement among candidate responses. At the dataset level, we apply per-dataset normalization to prevent datasets with disproportionately large reward scales or variances from dominating policy updates.
Together, these mechanisms transform routing optimization from directly fitting raw rewards into a calibrated learning process that explicitly separates objective-level supervision from distribution-level optimization variability. This allows the routing policy to focus on informative routing decisions while maintaining stable learning across heterogeneous tasks.

Our contributions are summarized as follows:
\begin{itemize}
    \item We propose a learning-signal calibration perspective for RL-based LLM routing, which improves policy learning by jointly modeling learning signal clarity and optimization comparability under heterogeneous tasks.
    \item We introduce a hierarchical and disentangled reward mechanism that performs component-wise advantage estimation before aggregation, enabling clearer credit assignment and reducing interference among competing objectives.
    \item We propose a distribution-aware optimization strategy that reduces optimization bias across heterogeneous data through variance-aware reweighting and per-dataset normalization.
    \item Extensive experiments on multiple datasets indicate that our method consistently improves routing performance and training stability.
\end{itemize}
\section{Related Work}

\subsection{LLM Routing and RL-based Optimization}
Existing LLM routing methods can be broadly categorized into three groups. Early approaches rely on heuristic or rule-based strategies~\cite{hu2024routerbench}, which use predefined criteria such as confidence or cost to select models. Supervised learning-based routers~\cite{ong2025RouteLLM,feng2024graphrouter,chen2024routerdc} further improve flexibility by training classifiers or scoring models to predict the best routing decisions. More recently, RL-based approaches optimize routing policies directly with reward signals, enabling adaptive trade-offs between performance and efficiency~\cite{Router-R1,chen2005qos}.
These methods typically optimize a scalar reward that combines multiple objectives, such as answer quality and format accuracy. While effective, these methods primarily focus on improving routing decisions themselves, with limited attention to how learning signal design affects policy optimization. 

\subsection{Learning Signal Design and Optimization}
Amount of RL-based work studies how to construct informative reward design, including reward shaping~\cite{ng1999policy}, multi-objective RL~\cite{hayes2022multi}, and advantage-based optimization~\cite{schulman2017ppo}. These approaches aim to improve how different aspects of feedback are represented and utilized during policy updates. However, these approaches mainly focus on improving the semantic quality of reward signals, rather than how heterogeneous reward distributions influence optimization behavior across instances and datasets.
Complementary to reward structuring, another line of work studies how heterogeneous data distributions affect optimization dynamics, focusing on techniques such as reward scaling and normalization in policy optimization~\cite{ouyang2022training,rafailov2023direct}, sample reweighting and off-policy correction~\cite{swaminathan2015counterfactual}, and curriculum or uncertainty-aware training~\cite{bengio2009curriculum,kendall2018multi}. These methods aim to stabilize training and improve data efficiency by adjusting the scale or importance of learning signals. However, they primarily study either the semantic structure of learning signals or their optimization behavior under heterogeneous data, but rarely consider the interaction between the two. This separation becomes limiting in LLM routing, where heterogeneous tasks jointly affect both what information learning signals contain and how strongly they influence policy updates.

\begin{figure*}[t]
    \centering
    \includegraphics[width=\textwidth]{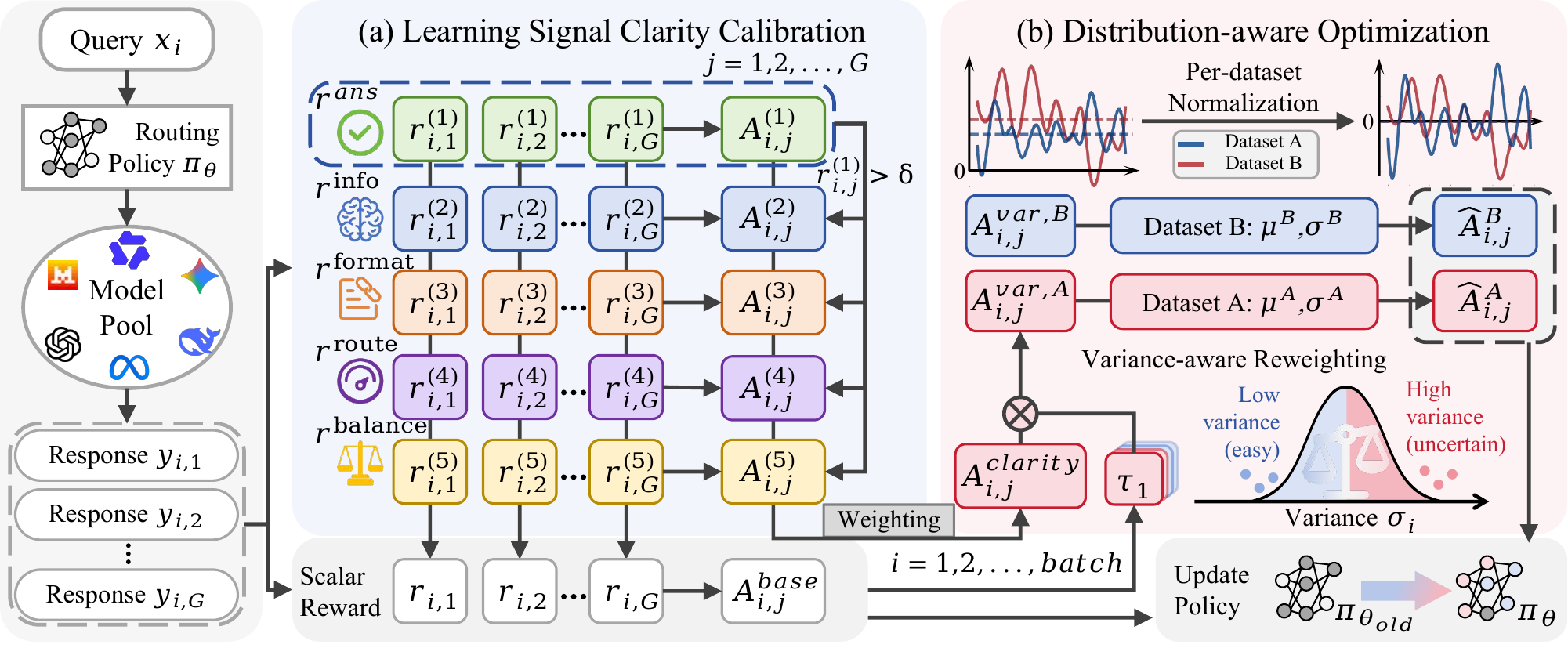}
    \caption{
    Overview of ReCal. Given a query, the routing policy samples multiple routing actions and obtains a response group. ReCal first decomposes routing rewards into multiple hierarchical objectives and performs component-wise advantage estimation before weighted aggregation. It then calibrates optimization variability through variance-aware reweighting and per-dataset normalization under heterogeneous reward distributions. 
    }
    \label{fig:framework}
    \vspace{-1em}
\end{figure*}

\section{Preliminaries}

\subsection{RL-based LLM Routing}

Given a query $x$, LLM routing selects an action $a$ from a candidate set $\mathcal{A}=\{a_1,\dots,a_K\}$ using a routing policy $\pi_\theta(a|x)$ parameterized by $\theta$, where $K$ denotes the number of routing candidates. After executing action $a \sim \pi_\theta(\cdot|x)$, the selected model generates a response $y$, which is evaluated by a reward function $r(x,a,y)$.

In practical routing systems, rewards are usually composed of multiple objectives, such as answer correctness, reasoning quality, formatting behavior, or routing efficiency. Existing methods typically aggregate them into a scalar reward:
\begin{equation}
r = \sum_{m=1}^{M} \lambda_m r^{(m)},
\end{equation}
where $M$ denotes the number of reward components, $r^{(m)}$ is the reward of the $m$-th objective, and $\lambda_m$ is its corresponding weight.

Given a training batch $\mathcal{D}=\{(x_i,a_i,y_i,r_i)\}_{i=1}^{N}$ with batch size $N$, RL-based routing optimizes the policy by maximizing the expected reward:
\begin{equation}
\max_\theta \ \mathbb{E}_{a \sim \pi_\theta(\cdot|x)}[r(x,a,y)].
\end{equation}

\subsection{Policy Optimization}

Most existing routing methods adopt Proximal Policy Optimization (PPO)-style~\cite{schulman2017ppo} optimization. Let $\pi_{\theta_{\text{old}}}$ denote the old routing policy. PPO optimizes the clipped objective $\mathcal{L}_{\text{PPO}}$:
\begin{equation}
\mathbb{E}
\left[
\min
\left(
\rho_t A_t,
\text{clip}(\rho_t,1-\epsilon,1+\epsilon)A_t
\right)
\right],
\end{equation}
where $\rho_t=\frac{\pi_\theta(a_t|x_t)}{\pi_{\theta_{\text{old}}}(a_t|x_t)}$ is the policy ratio for action $a_t$ under query $x_t$, $\epsilon$ is the clipping threshold, and $A_t$ denotes the estimated advantage, which is typically obtained directly from the scalarized reward, i.e., $A_t=\text{Adv}(r_t)$. 

Recent RL frameworks further adopt Group Relative Policy Optimization (GRPO)-style optimization. For each query $x_i$, the policy samples a response group $\mathcal{Y}_i=\{y_{i,1},\dots,y_{i,G}\}$ containing $G$ responses, where each response receives a scalar reward $r_{i,j}$. GRPO computes group-relative advantages by normalizing rewards within each group:
\begin{equation}
A_{i,j}
=
\frac{
r_{i,j}-\mu_i
}{
\sigma_i+\varepsilon
},
\end{equation}
where $\sigma_i=\sqrt{\frac{1}{G}\sum_{j=1}^{G}(r_{i,j}-\mu_i)^2}$ and $\mu_i=\frac{1}{G}\sum_{j=1}^{G}r_{i,j}$ denote the standard deviation and group mean, respectively, and $\varepsilon$ is a small constant for numerical stability.

\section{The Proposed Framework: ReCal}

Figure~\ref{fig:framework} illustrates the overall framework of ReCal. Starting from a response group sampled by the routing policy, ReCal performs a two-stage learning-signal calibration process before policy optimization. First, ReCal improves learning signal clarity through hierarchical reward decomposition and component-wise advantage estimation. It then calibrates optimization bias under heterogeneous reward distributions through variance-aware reweighting and per-dataset normalization. The resulting calibrated advantages are finally integrated into a GRPO-style policy optimization objective.

\subsection{Learning Signal Clarity Calibration}

Rather than estimating advantages directly from aggregated scalarized rewards, ReCal separates reward components before advantage estimation, allowing relative improvements under different objectives to be modeled independently.

\subsubsection{Hierarchical Reward Decomposition}

For each response $y_{i,j}$ in response group $\mathcal{Y}_i$, ReCal defines a set of reward components:
$
\mathbf{r}_{i,j}
=
\{
r_{i,j}^{\text{ans}},
r_{i,j}^{\text{info}},
r_{i,j}^{\text{format}},
r_{i,j}^{\text{route}},
r_{i,j}^{\text{balance}}
\}.
$
$r^{\text{ans}}$ is computed as the F1 score between the generated answer and the ground-truth. $r^{\text{info}}$ measures the quality of returned called-model information by computing the F1 score with the ground-truth. $r^{\text{format}}$ is a binary reward indicating whether the response follows the required output format. $r^{\text{route}}$ reflects routing efficiency based on the number of external model calls during reasoning. $r^{\text{balance}}$ encourages balanced router utilization by rewarding models with lower historical routing frequency.
To stabilize long-term routing behavior, the routing frequency statistics are accumulated with exponential moving average (EMA) decay. Let $c_t^{(k)}$ denote the accumulated routing count of model $a_k$ at training step $t$, and $\hat{c}_t^{(k)}$ denotes the routing count within the current step. The EMA-updated routing statistics are computed as:
\begin{equation}
c_t^{(k)}
=
\alpha c_{t-1}^{(k)}
+
\hat{c}_t^{(k)},
\end{equation}
where $\alpha \in [0,1)$ is the EMA decay coefficient. $r^{\text{balance}}$ is then inversely correlated with the normalized routing proportion of the selected model.

These objectives are organized hierarchically according to their dependency on answer correctness. Auxiliary rewards are activated only when the response satisfies a correctness threshold:
\begin{equation}
r_{i,j}^{(m)}
=
\mathbb{I}
(r_{i,j}^{\text{ans}}>\delta)
\cdot
\tilde{r}_{i,j}^{(m)},
\end{equation}
where $\tilde{r}_{i,j}^{(m)}$ denotes the raw reward, $\delta$ is the activation threshold, and $\mathbb{I}(\cdot)$ is the indicator function.

\subsubsection{Component-wise Advantage Estimation}

Instead of estimating advantages after scalar reward aggregation, ReCal computes group-relative advantages independently for each reward component. For the $m$-th reward objective, the component-wise advantage is computed as:
\begin{equation}
A_{i,j}^{(m)}
=
\frac{
r_{i,j}^{(m)}-\mu_i^{(m)}
}{
\sigma_i^{(m)}+\varepsilon
},
\end{equation}
where $\sigma_i^{(m)}
=
\sqrt{
\frac{1}{G}
\sum_{j=1}^{G}
(r_{i,j}^{(m)}-\mu_i^{(m)})^2
}$ and $\mu_i^{(m)}=\frac{1}{G}\sum_{j=1}^{G} r_{i,j}^{(m)}$ denote the standard deviation and mean of the $m$-th reward component within response group $\mathcal{Y}_i$.

The final clarity-calibrated advantage is then obtained through weighted aggregation:
\begin{equation}
A_{i,j}^{\text{clarity}}
=
\sum_{m=1}^{M}
\lambda_m
A_{i,j}^{(m)},
\end{equation}
where $\lambda_m$ denotes the weight of the $m$-th objective.

Compared with scalarized optimization, this formulation preserves objective-specific relative improvements before aggregation. As a result, the routing policy receives clearer, more discriminative supervision, enabling it to understand which samples are more conducive to policy improvement.

\subsection{Distribution-aware Optimization}

Although disentangled advantages improve learning signal clarity, optimization can still become biased under heterogeneous reward distributions. ReCal therefore further calibrates optimization variability across both instances and datasets.

\subsubsection{Variance-aware Reweighting}

For each response group $\mathcal{Y}_i$, with the reward mean $\mu_i^{\text{group}}=\frac{1}{G}\sum_{j=1}^{G}r_{i,j}$, we measures routing uncertainty using the reward variance within the group:
\begin{equation}
\sigma_i^{\text{group}}
=
\sqrt{
\frac{1}{G}
\sum_{j=1}^{G}
(r_{i,j}-\mu_i^{\text{group}})^2
}.
\end{equation}

Intuitively, larger intra-group variance indicates stronger disagreement among sampled routing decisions, suggesting that the corresponding query is more uncertain and potentially more informative.
ReCal therefore rescales optimization strength according to the relative variance level:
\begin{equation}
\tau_i
=
\text{clip}
\left(
\frac{
\sigma_i^{\text{group}}
}{
\bar{\sigma}
},
\tau_{\min},
\tau_{\max}
\right),
\end{equation}
where $\bar{\sigma}$ denotes the average group variance within the batch, and $\tau_{\min}$ and $\tau_{\max}$ are clipping thresholds.
Then the variance-calibrated advantage is computed as:
\begin{equation}
A_{i,j}^{\text{var}}
=
\tau_i
A_{i,j}^{\text{clarity}}.
\end{equation}

This mechanism allocates larger optimization weights to uncertain routing cases.

\subsubsection{Per-dataset Normalization}

Different datasets may exhibit substantially different reward magnitudes and variances, causing certain datasets to dominate optimization. To improve optimization comparability across heterogeneous data sources, ReCal normalizes advantages independently within each dataset.

Let $\mathcal{D}^{(d)}$ denote the set of samples from dataset $d$. We compute dataset-wise statistics:
\begin{equation}
\mu^{(d)}
=
\frac{1}{|\mathcal{D}^{(d)}|}
\sum_{(i,j)\in \mathcal{D}^{(d)}}
A_{i,j}^{\text{var}},
\end{equation}
\begin{equation}
\sigma^{(d)}
=
\sqrt{
\frac{1}{|\mathcal{D}^{(d)}|}
\sum_{(i,j)\in \mathcal{D}^{(d)}}
(A_{i,j}^{\text{var}}-\mu^{(d)})^2
}.
\end{equation}

The final calibrated advantage is:
\begin{equation}
\hat{A}_{i,j}
=
\frac{
A_{i,j}^{\text{var}}-\mu^{(d)}
}{
\sigma^{(d)}+\epsilon
}.
\end{equation}

This normalization aligns optimization scales across datasets and reduces optimization bias caused by heterogeneous reward distributions.

\subsection{Overall Training Objective}

$\hat{A}_{i,j}$ is integrated into a GRPO-style clipped policy optimization objective $\mathcal{L}_{\text{ReCal}}$:
\begin{equation}
\mathbb{E}
\left[
\min
\left(
\rho_{i,j}\hat{A}_{i,j},
\text{clip}
(\rho_{i,j},1-\epsilon,1+\epsilon)
\hat{A}_{i,j}
\right)
\right],
\end{equation}
where
$
\rho_{i,j}
=
\frac{
\pi_\theta(a_{i,j}|x_i)
}{
\pi_{\theta_{\text{old}}}(a_{i,j}|x_i)
}
$
denotes the policy ratio between the updated policy and the previous policy.

ReCal explicitly calibrates both learning signal clarity and optimization weighting before policy updates, resulting in more stable and informative policy learning under heterogeneous scenarios.
\begin{table*}[t]
    \centering
\setlength{\tabcolsep}{8pt}
\resizebox{0.95\textwidth}{!}{
    \begin{tabular}{lcccccccc}
        \toprule
        \multirow{2}{*}{\textbf{Methods}} & \multicolumn{3}{c}{\textbf{General QA}} & \multicolumn{4}{c}{\textbf{Multi-Hop QA}} & \multirow{2}{*}{\textbf{Avg.}}\\
        \cmidrule{2-4} \cmidrule{5-8}
         & \textbf{NQ$^\dagger$} & \textbf{TriviaQA} & \textbf{PopQA} & \textbf{HpQA$^\dagger$} & \textbf{2wiki} & \textbf{Musique} & \textbf{Bamb}\\
        \midrule
        \rowcolor{white}
        \rowcolor{softred}\multicolumn{9}{c}{\textbf{\textit{No Routing}}} \\
        Vanilla & 0.092 & 0.260 & 0.122 & 0.140 & 0.266 & 0.026 & 0.040 & 0.135 \\
        CoT & 0.126 & 0.358 & 0.160 & 0.168 & 0.208 & 0.046 & 0.224 & 0.184 \\
        SFT & 0.212 & 0.400 & 0.160 & 0.198 & 0.256 & 0.052 & 0.112 & 0.199 \\
        RAG & 0.298 & 0.540 & 0.366 & 0.216 & 0.146 & 0.078 & 0.224 & 0.267 \\
        Search-R1 & 0.328 & 0.510 & 0.324 & 0.236 & 0.278 & 0.090 & 0.272 & 0.291 \\
        \midrule
        \rowcolor{table-yellow}\multicolumn{9}{c}{\textbf{\textit{Heuristic \& Discriminative Routing}}} \\
        Largest LLM & 0.296 & 0.578 & 0.354 & 0.278 & 0.274 & 0.104 & 0.480 & 0.338 \\
        Prompt LLM & 0.300 & 0.580 & 0.340 & 0.268 & 0.262 & 0.108 & 0.448 & 0.329 \\
        Prompt LLM* & 0.258 & 0.500 & 0.256 & 0.206 & 0.248 & 0.078 & 0.472 & 0.288 \\
        KNN Router & 0.262 & 0.528 & 0.222 & 0.224 & 0.196 & 0.066 & 0.360 & 0.265 \\
        KNN Router* & 0.236 & 0.478 & 0.232 & 0.154 & 0.234 & 0.072 & 0.384 & 0.256 \\
        MLP Router & 0.252 & 0.460 & 0.222 & 0.198 & 0.210 & 0.072 & 0.360 & 0.253 \\
        RouteLLM & 0.230 & 0.516 & 0.192 & 0.216 & 0.206 & 0.058 & 0.312 & 0.247 \\
        RouterDC & 0.278 & 0.592 & 0.282 & 0.244 & 0.218 & 0.080 & 0.504 & 0.314 \\
        GraphRouter & 0.276 & 0.586 & 0.280 & 0.234 & 0.180 & 0.076 & 0.448 & 0.297 \\
        \midrule
        \rowcolor{table-yellow}\multicolumn{9}{c}{\textbf{\textit{RL-based Routing}}} \\
        Router-R1 & 0.334 & 0.668 & 0.400 & 0.412 & 0.476 & 0.186 & 0.424 & 0.414 \\
        \rowcolor{table-blue}\textbf{Ours} & \textbf{0.404} & \textbf{0.712} & \underline{0.466} & \underline{0.442} & \textbf{0.560} & \textbf{0.212} & \underline{0.568} & \textbf{0.481} \\
        \rowcolor{table-blue}\textbf{Ours+} & \underline{0.390} & \underline{0.698} & \textbf{0.466} & \textbf{0.448} & \underline{0.554} & \underline{0.206} & \textbf{0.584} & \underline{0.471} \\
        \bottomrule
    \end{tabular}}
    \caption{Experimental results on seven QA datasets w.r.t. Exact Match. \textbf{Bold} indicates the best score in each column for each method, and \underline{underline} indicates the secondary.\small $^\dagger$ indicates in-domain evaluation; all others are out-of-domain.}
    \label{tab:exp_main}
    \vspace{-0.8em}
\end{table*}
\section{Experiments}
\subsection{Experimental Setup}
\paragraph{Datasets}
We evaluate ReCal on seven question-answering (QA) datasets: (1) General QA: Natural Question (NQ)~\cite{natural-questions}, TriviaQA~\cite{joshi-etal-2017-triviaqa}, PopQA~\cite{popqa}; (2) Multi-Hop QA: HotpotQA (HpQA)~\cite{yang2018hotpotqa}, 2WikiMultiHopQA (2wiki)~\cite{2wikimultihopqa}, Musique~\cite{trivedi-etal-2022-musique}, and Bamboogle (Bamb)~\cite{bamboogle}.

\paragraph{Baselines}
We compare ReCal with  three categories of methods: \textbf{(1) No routing}: 
Direct Inference (Vanilla), Chain-of-Thought (CoT)~\cite{wei2022cot}, supervised
fine-tuning (SFT), Retrieval-Augmented Generation (RAG) based on the external knowledge Wikipedia-18~\cite{wikipedia} and the retriever E5~\cite{wang2024E5}, and Search-R1~\cite{jin2025searchr1}; 
\textbf{(2) Heuristic \& Discriminative routing}: 
Largest LLM, Prompt LLM, Prompt LLM* (Prompt LLM + task decomposition),
KNN Router\cite{hu2024routerbench}, KNN Router* (KNN Router + task decomposition), MLP Router\cite{hu2024routerbench}, RouteLLM\cite{ong2025RouteLLM}, RouterDC\cite{chen2024routerdc}, and GraphRouter\cite{feng2024graphrouter}; \textbf{(3) RL-based routing}: Router-R1\cite{Router-R1}.

\paragraph{Implementation Details}
We used Qwen2.5-3B-instruct~\cite{qwen2025qwen25} as the base model and conducted RL training using GRPO within the VeRL~\footnote{\url{https://github.com/verl-project/verl}} framework on NVIDIA A100 GPUs, with a batch size of 64 and up to 150 training steps. Similar to Router-R1, we combined 7K samples each from the NQ and HotpotQA datasets to create a training set. After training, we evaluated performance on the two in-domain datasets and the other five out-of-domain datasets by sampling 500 test instances from each of them (excluding Bamboogle, which contained a total of 125 test instances). All baseline models were trained (where applicable) and evaluated on consistent datasets and settings. We selected a set of six models of varying sizes and architectures via the NVIDIA NIM API~\footnote{\url{https://build.nvidia.com/}} and OpenRouter~\footnote{\url{https://openrouter.ai/}}: Qwen2.5-7B-Instruct~\cite{qwen2025qwen25}, LLaMA-3.1-8B-Instruct~\cite{grattafiori2024llama3}, LLaMA-3.1-70B-Instruct~\cite{grattafiori2024llama3}, Mistral-7B-Instruct~\cite{jiang2023identify}, Mixtral8x22B-Instruct~\cite{jiang2024mixtralexperts}, and Gemma-2-27B-Instruct~\cite{gemma2024gemma2}.
\begin{table*}[t]
    \centering
    \rowcolors{1}{white}{gray!10}
\setlength{\tabcolsep}{10pt}
\resizebox{0.9\textwidth}{!}{
\begin{tabular}{ccccccccc}
        \toprule
        \rowcolor{white}
        \textbf{Methods} & \textbf{NQ$^\dagger$} & \textbf{TriviaQA} & \textbf{PopQA} & \textbf{HpQA$^\dagger$} & \textbf{2wiki} & \textbf{Musique} & \textbf{Bamb} & \textbf{Avg.} \\
        \midrule
w/o both                     & 0.324          & 0.618          & 0.332          & 0.354          & 0.334          & 0.132          & 0.456          & 0.364          \\
w/o LSC                      & 0.346          & 0.666          & 0.390          & 0.378          & 0.464          & 0.156          & 0.528          & 0.418          \\
w/o CAE                       & 0.330          & 0.662          & 0.420          & 0.408          & 0.496          & 0.160          & 0.536          & 0.430          \\
w/o DAO                      & \underline{0.366}    & \underline{0.690}    & \underline{0.438}    & \underline{0.436}    & \underline{0.502}    & \underline{0.188}    & \underline{0.568}    & \underline{0.456}    \\
\rowcolor{table-blue}Ours                         
& \textbf{0.404} & \textbf{0.712} & \textbf{0.466} & \textbf{0.442} & \textbf{0.560} & \textbf{0.212} & \textbf{0.568} & \textbf{0.481}\\
\bottomrule
\end{tabular}}
    \caption{Results of Ablation. LSC means \textit{\textbf{Learning Signal Clarity}} Calibration, CAE means \textit{\textbf{Component-wise Advantage Estimation}}, DAO means \textit{\textbf{Distribution-aware Optimization}}. $^\dagger$ indicates in-domain evaluation.}
    \label{tab:abla}
    \vspace{-1em}
\end{table*}
\begin{figure*}[ht]
    \centering
    \includegraphics[width=0.86\textwidth]{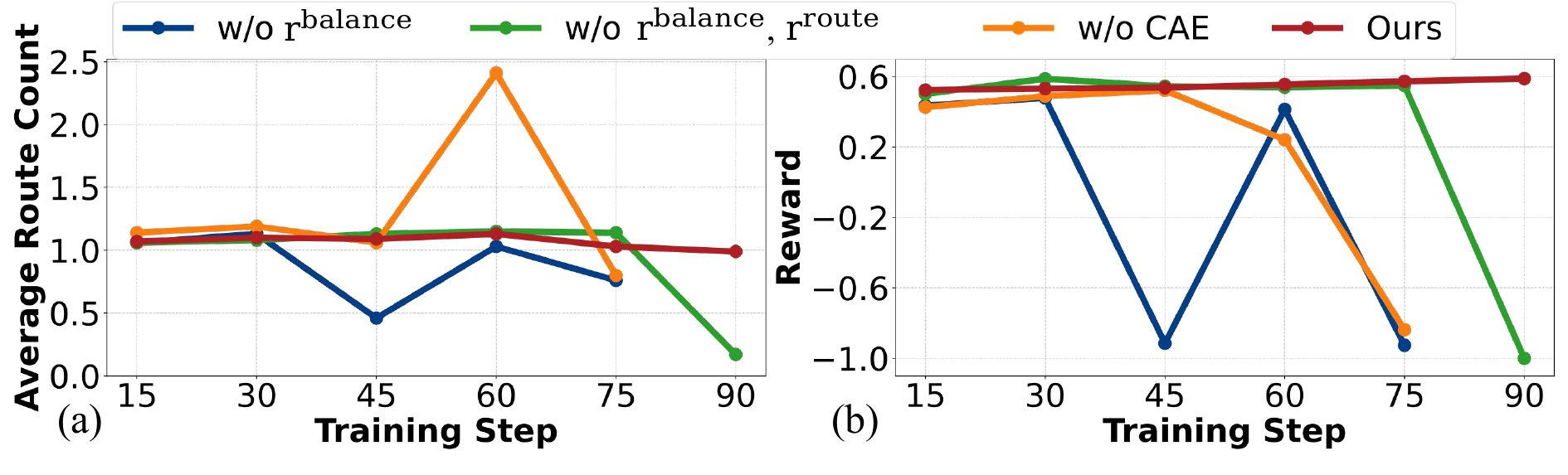}
    \vspace{-1em}
    \caption{As the training steps increase, (a) and (b) display the average route count and reward for Ours, and for versions that remove $r^{balance}$, $r^{balance}\&r^{route}$, and CAE, respectively.}
    \label{fig:ana0}
    \vspace{-1em}
\end{figure*}
\subsection{Main Results}

\begin{figure*}[ht]
    \centering
    \includegraphics[width=0.88\textwidth]{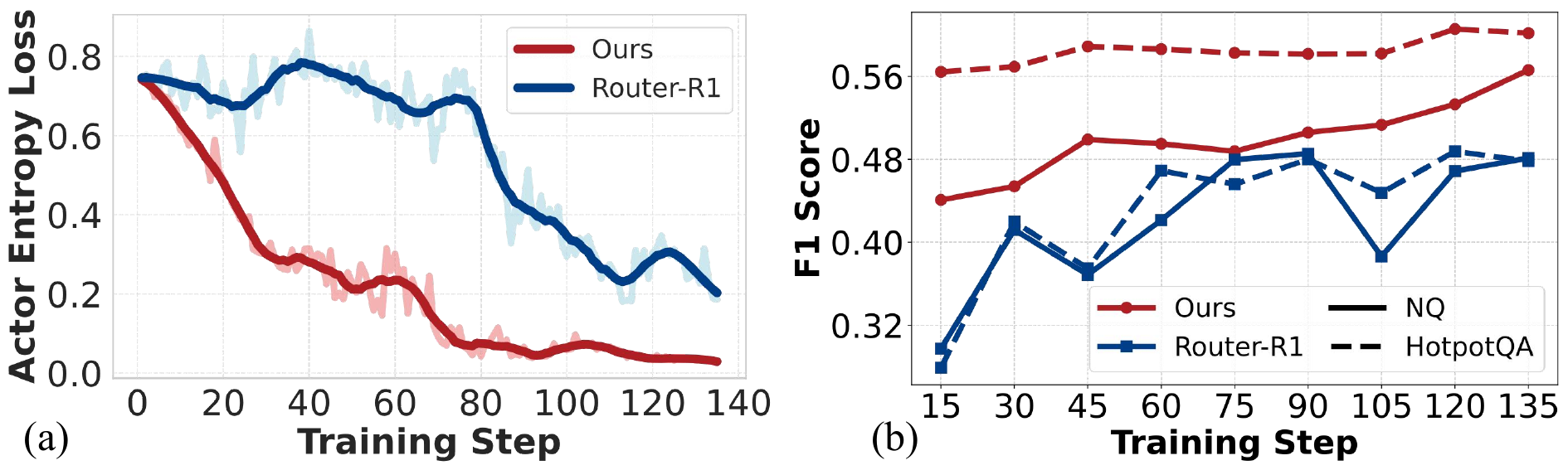}
    \caption{
    As the training steps increase, (a) shows the entropy loss convergence curves for Ours and Router-R1. (b) shows the F1 scores of Ours and Router-R1 on the NQ and HotpotQA test sets.}
    \label{fig:ana1}
    \vspace{-1em}
\end{figure*}
As shown in Table~\ref{tab:exp_main},  our proposed ReCal achieves the best average performance among all compared methods on seven QA benchmarks, demonstrating the effectiveness of learning signal calibration for RL-based LLM routing.
Compared with non-routing methods, RL-based routing substantially improves performance across all QA tasks, validating the importance of adaptive model selection. Router-R1 achieves strong gains over heuristic and discriminative routers, showing the advantage of end-to-end policy optimization. However, our method consistently outperforms Router-R1 on most datasets, indicating that improving the clarity and comparability of learning signals is critical for effective routing optimization.

\textbf{Ours} achieves the best overall average EM score of $0.481$, surpassing Router-R1 by $6.7\%$ relatively ($0.481$ vs.\ $0.414$). The improvements are particularly significant on more challenging multi-hop QA benchmarks. 
For example, ReCal improves performance from $0.476$ to $0.560$ on 2Wiki and from $0.424$ to $0.568$ on Bamboogle.
These results suggest that calibrating learning signals is especially beneficial for routing policy with high uncertainty.

On in-domain datasets, ReCal also demonstrates strong improvements. Compared with Router-R1, our method improves EM from $0.334$ to $0.404$ on NQ and from $0.412$ to $0.442$ on HotpotQA. More importantly, the gains consistently transfer to unseen out-of-domain datasets, indicating that the proposed calibration mechanisms improve the generalization ability of routing policies rather than overfitting to specific training distributions.

We further observe that heuristic and discriminative routing methods exhibit relatively unstable performance across datasets. Prompt LLM and RouterDC perform competitively on certain datasets, such as TriviaQA or Bamboogle, but their overall performance remains substantially lower than RL-based methods. This suggests that heuristics or discriminative routing strategies struggle to adapt to heterogeneous requirements across tasks.

\subsection{In-depth Analysis}

\subsubsection{Ablation Study}

As shown in Table~\ref{tab:abla}, removing either Learning Signal Clarity Calibration \textbf{(LSC)} or Distribution-aware Optimization \textbf{(DAO)} consistently degrades performance across datasets, while removing both causes the largest drop. This verifies that both learning signal clarity and optimization comparability are important for RL-based routing.

Removing LSC decreases the average score to $0.418$, indicating that directly optimizing scalarized rewards leads to weaker credit assignment and noisier optimization signals. Removing only Component-wise Advantage Estimation \textbf{(CAE)} still causes a noticeable drop to $0.430$, demonstrating that component-wise advantage estimation itself is critical for preserving objective-specific supervision during optimization.
DAO further improves performance from $0.456$ to $0.481$, with especially clear gains on challenging multi-hop datasets such as 2Wiki and Musique. This suggests that heterogeneous reward distributions introduce optimization bias during RL training, and distribution-aware calibration helps stabilize learning across different datasets and routing cases.

As Figure~\ref{fig:ana0} shows, removing $r^{\text{balance}}$ introduces obvious training instability and leads to a substantial decrease in reward during the optimization, indicating that the balance regularization term is crucial for preventing the routing policy from collapsing. Removing both $r^{\text{balance}}$ and $r^{\text{route}}$ also sharply decreases both the routing counts and the reward in the later phases of training.
Removing CAE produces a different failure mode: the routing count exhibits abnormal spikes while the reward rapidly collapses later. This indicates that directly estimating advantages from scalarized rewards introduces entangled and unstable optimization signals. In contrast, the full ReCal framework maintains the smoothest routing behavior and the most stable reward improvement throughout training.

\subsubsection{Stability and Convergence of Training}
Figure~\ref{fig:ana1}(a) presents the entropy loss curves during RL training. ReCal exhibits substantially smoother and faster convergence, with entropy steadily decreasing throughout training and converging to a low-variance regime in early stages. In contrast, Router-R1 shows persistent oscillations and slower convergence, indicating unstable policy updates under scalarized reward optimization.

As shown in Figure~\ref{fig:ana1}(b), ReCal consistently yields more stable performance gains across both datasets, whereas Router R1 exhibited significant fluctuations and competition across the datasets. In particular, during training steps $60$--$75$, Router-R1 improves on NQ while simultaneously degrading on HotpotQA, suggesting that the optimization process becomes biased toward dominant dataset-specific reward signals. This behavior reveals the existence of cross-dataset optimization interference under heterogeneous reward distributions.
By contrast, ReCal jointly improves performance on both datasets throughout training. These results demonstrate that calibrating both reward clarity and optimization distribution improves not only routing performance but also the stability and consistency of policy optimization under heterogeneous tasks.

\subsubsection{Generalization to Unseen LLMs}
To evaluate the generalization ability of ReCal under expanding model pools, we introduce two additional unseen LLMs, DeepSeek-V3~\cite{deepseekai2025deepseekv3} and Qwen2.5-72B-Instruct~\cite{qwen2025qwen25}, during inference only. We augment the routing prompt with textual descriptions of the newly models, allowing the routing policy to select them.
As shown in Table~\ref{tab:exp_main}, \textbf{Ours+} maintains strong performance and further improves several datasets. This result suggests that ReCal does not overfit to a fixed routing space. Instead, the learned routing policy can generalize to newly introduced models by leveraging semantic descriptions.

\section{Conclusion}
In this paper, we propose ReCal, a reward calibration framework for RL-based LLM routing. ReCal introduce a hierarchical reward decomposition mechanism with component-wise advantage estimation to provide clearer objective-level supervision, together with a distribution-aware optimization strategy that calibrates optimization variability across heterogeneous instances and datasets through variance-aware reweighting and per-dataset normalization. Extensive experiments demonstrate that ReCal consistently improves routing performance, training stability, and generalization over routing baselines. Further analyses show that ReCal produces more stable optimization under heterogeneous routing scenarios.
Beyond the current setting, we believe ReCal provides a promising direction for future RL-based LLM routing. In future work, we plan to extend ReCal to broader routing scenarios. Another important direction is to develop more adaptive and automatic calibration strategies that can dynamically adjust reward structures and optimization variability under evolving model pools and task distributions.

\newpage
\section*{Limitations}
ReCal is evaluated primarily on QA-oriented routing tasks, and its effectiveness for other routing settings, such as multimodal systems remains to be explored. Furthermore, ReCal currently relies on manually designed reward components and weighting coefficients, which may require additional tuning when adapting to new tasks or model pools.
\section*{Ethical Considerations}
We use only publicly available datasets that are widely adopted in the LLM community and contain no personally identifiable information (PII) in our experiments. All evaluated models are accessed through publicly available APIs or open-source checkpoints under their respective licenses.

\bibliography{emnlp26}

\appendix

\section{Case Study}
\begin{tcolorbox}[colback=red!25!green!30!blue!5!white,colframe=red!20!green!40!blue,title={
Response of \textbf{ReCal}}]

\think{To answer the question ""Who is responsible for introducing the principle of uniformitarianism?"", I need to perform a search for the historical context and the originator of this geological principle. I will seek information from a specialized LLM to accurately identify the person who introduced this principle.}

\route{LLaMA-3.1-70B-Instruct: Who is responsible for introducing the principle of uniformitarianism?}

\info{The principle of uniformitarianism is attributed to James Hutton (1726-1797), a Scottish geologist and physician. However, it was Charles Lyell (1797-1875), an English geologist, who popularized and further developed this concept in his influential book ""Principles of Geology"" (1830-1833). Lyell's work built upon Hutton's ideas and presented a comprehensive framework for understanding the Earth's geological history through the principle of uniformitarianism.
In essence, while Hutton introduced the concept, Lyell is often credited with fully articulating and promoting the principle of uniformitarianism, making it a cornerstone of modern geology.}
\end{tcolorbox}
\begin{tcolorbox}[colback=red!25!green!30!blue!5!white,colframe=red!20!green!40!blue]

\think{After consulting the information provided, I now have a clearer understanding of the role of James Hutton and Charles Lyell in the development of the principle of uniformitarianism. Hutton is credited with the initial introduction of the concept, but Lyell is recognized for popularizing and fully establishing this principle in geological studies.}

\route{Mixtral-8x22B-Instruct: Who is responsible for introducing the principle of uniformitarianism?}

\info{The principle of uniformitarianism was introduced by the Scottish geologist James Hutton. He proposed this principle in the late 18th century, suggesting that the geological processes that shaped Earth in the past are the same slow-acting processes that we observe today. This principle contrasted with catastrophism, which held that Earth's features were formed by sudden, violent events. Hutton's uniformitarianism, also known as the theory of gradualism, had a significant impact on the development of modern geology.}

\answer{James Hutton}
\end{tcolorbox}
\begin{tcolorbox}[colback=red!25!green!30!blue!5!white,colframe=red!20!green!40!blue,title={Response of \textbf{Router-R1}}]

\think{The principle of uniformitarianism is a theory of the earth’s past and present processes, so historically it was more commonly developed through empirical observations and integrative analysis rather than solely credited to a single person. Multiple researchers often make significant contributions as they gather, analyze, or theorize about the same evidence, sometimes reinforcing and building on each other’s work.}

\route{LLaMA-3.1-70B-Instruct: Who introduced the principle of uniformitarianism? When answering, consider historical evidence and recognition among scientists for the principle's development.}
\end{tcolorbox}

\begin{table*}[t]
    \centering
\setlength{\tabcolsep}{8.5pt}
\resizebox{0.92\textwidth}{!}{
    \begin{tabular}{lcccccccc}
        \toprule
        \multirow{2}{*}{\textbf{Methods}} & \multicolumn{3}{c}{\textbf{General QA}} & \multicolumn{4}{c}{\textbf{Multi-Hop QA}} & \multirow{2}{*}{\textbf{Avg.}}\\
        \cmidrule{2-4} \cmidrule{5-8}
         & \textbf{NQ$^\dagger$} & \textbf{TriviaQA} & \textbf{PopQA} & \textbf{HpQA$^\dagger$} & \textbf{2wiki} & \textbf{Musique} & \textbf{Bamb}\\
        \midrule
        \rowcolor{white}
        \rowcolor{softred}\multicolumn{9}{c}{\textbf{\textit{No Routing}}} \\
        Vanilla & 0.162 & 0.341 & 0.154 & 0.215 & 0.304 & 0.081 & 0.112 & 0.196 \\
        CoT & 0.218 & 0.431 & 0.185 & 0.260 & 0.251 & 0.106 & 0.332 & 0.255 \\
        SFT & 0.289 & 0.460 & 0.207 & 0.281 & 0.291 & 0.121 & 0.173 & 0.260 \\
        RAG & 0.414 & 0.622 & 0.452 & 0.307 & 0.187 & 0.134 & 0.303 & 0.346 \\
        Search-R1 & 0.407 & 0.575 & 0.383 & 0.328 & 0.317 & 0.145 & 0.387 & 0.363 \\
        \midrule
        \rowcolor{table-yellow}\multicolumn{9}{c}{\textbf{\textit{Heuristic \& Discriminative Routing}}} \\
        Largest LLM & 0.431 & 0.695 & 0.423 & 0.416 & 0.397 & 0.199 & 0.608 & 0.453 \\
        Prompt LLM & 0.437 & 0.694 & 0.409 & 0.407 & 0.393 & 0.205 & 0.579 & 0.446 \\
        Prompt LLM* & 0.373 & 0.600 & 0.315 & 0.313 & 0.355 & 0.165 & 0.580 & 0.386 \\
        KNN Router & 0.388 & 0.627 & 0.281 & 0.341 & 0.289 & 0.141 & 0.496 & 0.366 \\
        KNN Router* & 0.360 & 0.572 & 0.272 & 0.248 & 0.299 & 0.138 & 0.493 & 0.340 \\
        MLP Router & 0.368 & 0.557 & 0.272 & 0.295 & 0.277 & 0.142 & 0.460 & 0.339 \\
        RouteLLM & 0.362 & 0.635 & 0.251 & 0.335 & 0.284 & 0.135 & 0.423 & 0.346 \\
        RouterDC & 0.410 & 0.694 & 0.328 & 0.381 & 0.293 & 0.165 & 0.623 & 0.413 \\
        GraphRouter & 0.412 & 0.690 & 0.324 & 0.366 & 0.258 & 0.154 & 0.546 & 0.393 \\
        \midrule
        \rowcolor{table-yellow}\multicolumn{9}{c}{\textbf{\textit{RL-based Routing}}} \\
        Router-R1 & 0.449 & 0.745 & 0.458 & 0.494 & 0.529 & 0.263 & 0.596 & 0.504 \\
        \rowcolor{table-blue}\textbf{Ours} & \textbf{0.508} & \textbf{0.784} & \textbf{0.505} & \underline{0.529}    & \underline{0.613}    & \textbf{0.316} & \underline{0.690}    & \textbf{0.564} \\
        \rowcolor{table-blue}\textbf{Ours+} & \underline{0.499}    & \underline{0.776}    & \underline{0.502}    & \textbf{0.534} & \textbf{0.613} & \underline{0.308}    & \textbf{0.710} & \underline{0.563}   \\
        \bottomrule
    \end{tabular}}
    \caption{Experimental results on seven QA datasets w.r.t. F1 score. \textbf{Bold} indicates the best score in each column for each method, and \underline{underline} indicates the secondary.\small $^\dagger$ indicates in-domain evaluation; all others are out-of-domain.}
    \label{tab:f1}
    \vspace{-1em}
\end{table*}
For the question \textbf{\textit{"Who is responsible for introducing the principle of uniformitarianism?"}}, the correct answer is \textbf{\textit{['James Hutton' 'geologist James Hutton']}}. Above are the responses generated by Ours (ReCal) and Router-R1. Our approach ensures the correctness of the answer by invoking different models, whereas Router-R1 easily provides incorrect answers in complex contexts.
\begin{tcolorbox}[colback=red!25!green!30!blue!5!white,colframe=red!20!green!40!blue]

\info{The principle of uniformitarianism was introduced by James Hutton (1726-1797) and further developed by Charles Lyell (1797-1875). However, it was Lyell who popularized and extensively applied the concept in his influential book ""Principles of Geology"" (1830-1833).
Lyell's work built upon Hutton's earlier ideas, and his book became a seminal text in the field of geology. Lyell's uniformitarianism was widely accepted by scientists, including Charles Darwin, who was heavily influenced by Lyell's work during his voyage on the Beagle.
Historical evidence and recognition among scientists confirm that Lyell is often credited with the development and popularization of uniformitarianism, while Hutton is acknowledged as a precursor to Lyell's work.}

\answer{Charles Lyell}
\end{tcolorbox}

\section{More Experimental Results}
We present extensive experimental results on seven QA datasets with respect to F1-Score in Table~\ref{tab:f1}. ReCal consistently outperforms all baselines across both general and multi-hop QA tasks, achieving the highest average F1-scores. It maintains high accuracy even after introducing unseen models, and achieves state-of-the-art performance.
\section{Hyperparameter Settings}

\begin{table}[htbp]
    \centering
    \begin{tabularx}{\linewidth}{lX}
        \toprule
        \textbf{Hyperparameter} & \textbf{Value}\\
        \midrule
        Learning Rate (Actor) & 1e-6 \\
        Total Batch Size & 64 \\
        Max Training Steps & 150 \\
        Max Routing Steps & 3 \\
        Response group size $G$ & 4 \\
        GPU Utilization Ratio & 0.6 \\
        Rollout Sampling Temperature & 1.0 \\
        EMA decay $\alpha$ & 0.9 \\
$\tau_{\min}, \tau_{\max}$ & 0.8, 1.25 \\
        \bottomrule
    \end{tabularx}
    \caption{Hyperparameter Settings}
\end{table}

\section{Notation}
\begin{table*}[t]
\centering
\resizebox{\textwidth}{!}{
\begin{tabular}{p{3.8cm}p{11.8cm}}
\toprule
\multicolumn{2}{c}{\textbf{Notation Table}} \\
\midrule

\rowcolor{gray!15}
\multicolumn{2}{l}{\textbf{(A) Routing and Policy Optimization}} \\

$x$ & Input query. \\

$\mathcal{A}=\{a_1,\dots,a_K\}$ 
& Candidate routing action set, where $K$ is the number of candidates. \\

$a$ 
& Routing action sampled from the routing policy. \\

$\pi_\theta(a|x)$ 
& Routing policy parameterized by $\theta$. \\

$\theta$ 
& Parameters of the routing policy. \\

$y$ 
& Response generated after selecting routing action $a$. \\

$\mathcal{D}=\{(x_i,a_i,y_i,r_i)\}_{i=1}^{N}$ 
& Training batch with batch size $N$. \\

$
\rho_{i,j}
=
\frac{
\pi_\theta(a_{i,j}|x_i)
}{
\pi_{\theta_{\text{old}}}(a_{i,j}|x_i)
}
$
& PPO/GRPO probability ratio between updated and old policies. \\

$\epsilon$ 
& Clipping threshold in PPO/GRPO optimization. \\

$\varepsilon$ 
& A small constant for numerical stability. \\

$\mathcal{L}_{\text{ReCal}}$ 
& Final GRPO-style training objective of ReCal. \\

\midrule

\rowcolor{gray!15}
\multicolumn{2}{l}{\textbf{(B) Response Group and Reward Modeling}} \\

$\mathcal{Y}_i=\{y_{i,1},\dots,y_{i,G}\}$ 
& A group of $G$ responses sampled for query $x_i$. \\

$r(x,a,y)$ 
& Scalar reward evaluating the routing decision and generated response. \\

$r^{(m)}$ 
& Reward component of the $m$-th objective. \\

$M$ 
& Number of reward components. \\

$\lambda_m$ 
& Weight assigned to the $m$-th reward component. \\

$\mathbf{r}_{i,j}$ 
& Reward vector of response $y_{i,j}$. \\

$r_{i,j}^{\text{ans}}$ 
& Answer correctness reward based on F1 between prediction and ground-truth. \\

$r_{i,j}^{\text{info}}$ 
& Information-quality reward computed from the F1 between generated supporting information and ground-truth. \\

$r_{i,j}^{\text{format}}$ 
& Formatting reward indicating whether the response satisfies the format. \\

$r_{i,j}^{\text{route}}$ 
& Routing-efficiency reward related to external model invocation frequency. \\

$r_{i,j}^{\text{balance}}$ 
& Routing-balance reward encouraging balanced model utilization. \\

$\tilde{r}_{i,j}^{(m)}$ 
& Raw auxiliary reward before hierarchical gating. \\

$\delta$ 
& Correctness threshold for activating auxiliary rewards. \\

$\mathbb{I}(\cdot)$ 
& Indicator function. \\

\midrule

\rowcolor{gray!15}
\multicolumn{2}{l}{\textbf{(C) Advantage Estimation and Calibration}} \\

$A_t$ 
& Standard advantage estimate used in PPO-style optimization. \\

$A_{i,j}^{(m)}$ 
& Component-wise advantage for the $m$-th reward objective. \\

$A_{i,j}^{\text{clarity}}$ 
& Aggregated advantage after component-wise advantage estimation. \\

$A_{i,j}^{\text{var}}$ 
& Variance-aware reweighted advantage. \\

$\hat{A}_{i,j}$ 
& Final calibrated advantage after dataset-level normalization. \\

$\mu_i^{(m)}$ 
& Mean reward of the $m$-th component within response group $\mathcal{Y}_i$. \\

$\sigma_i^{(m)}$ 
& Standard deviation of the $m$-th reward component within response group $\mathcal{Y}_i$. \\

$\mu_i$ 
& Mean scalar reward within response group $\mathcal{Y}_i$. \\

$\sigma_i^{\text{group}}$ 
& Intra-group reward standard deviation used to measure routing uncertainty. \\

$\bar{\sigma}$ 
& Average intra-group standard deviation within the current batch. \\

$\tau_i$ 
& Variance-aware scaling factor for query $x_i$. \\

$\tau_{\min},\tau_{\max}$ 
& Lower and upper clipping bounds for $\tau_i$. \\

\midrule

\rowcolor{gray!15}
\multicolumn{2}{l}{\textbf{(D) Dataset-level Normalization}} \\

$\mathcal{D}^{(d)}$ 
& Set of samples belonging to dataset $d$. \\

$\mu^{(d)}$ 
& Mean advantage value within dataset $d$. \\

$\sigma^{(d)}$ 
& Standard deviation of advantages within dataset $d$. \\

$d$ 
& Dataset index. \\

\bottomrule
\end{tabular}}
\caption{Summary of notations used in ReCal.}
\label{tab:notation}
\end{table*}

\end{document}